\documentclass[journal]{IEEEtran}
\usepackage{amsmath,amsfonts}
\usepackage{algorithmic}
\usepackage{array}
\usepackage[sort&compress,numbers]{natbib}
\usepackage{textcomp}
\usepackage{stfloats}
\usepackage{subfloat}
\usepackage{subfigure}

\usepackage{color}

\usepackage{bbding}

\usepackage{algorithm}
\usepackage{algorithmic}

\usepackage{color}

\usepackage{url}

\usepackage{multirow}
\usepackage{verbatim}
\usepackage{graphicx}

\usepackage[table,xcdraw]{xcolor}

\usepackage{float}
\usepackage{booktabs}

\ifCLASSINFOpdf
\else
\fi
\begin{document}
%
\title{Seeing is not Believing: An Identity Hider for  Human Vision Privacy Protection}
%
%
%

	\author{Tao Wang, Yushu Zhang, Zixuan Yang, Xiangli Xiao, Hua Zhang, and Zhongyun Hua

	\thanks{
		
	This work was supported  by the Guangdong Provincial Key	Laboratory of Novel Security Intelligence Technologies under Grant No. 2022B1212010005. \textit{(Corresponding author: Yushu Zhang)}

		Tao Wang, Yushu Zhang, Zixuan Yang, and Xiangli Xiao are with the College of Computer Science and Technology, Nanjing University of Aeronautics and Astronautics, Nanjing 211106, China, and also with the Guangdong Provincial Key Laboratory of
		Novel Security Intelligence Technologies, Shenzhen 518055, China (e-mail: \{wangtao21; yushu; yangzixuan; xiaoxiangli\}@nuaa.edu.cn).

		Hua Zhang is with the State Key Laboratory of Information Security,
		Institute of Information Engineering, Chinese Academy of Sciences, Beijing
		100093, China (e-mail: zhanghua@iie.ac.cn).

		Zhongyun Hua is with the School of Computer Science and Technology, Harbin Institute of Technology (Shenzhen), Shenzhen 518055, China, and also with the Guangdong Provincial Key Laboratory of Novel Security Intelligence Technologies, Shenzhen, 518055, China (e-mail: huazhongyun@hit.edu.cn).

}}
%
%

\markboth{Journal of \LaTeX\ Class Files,~Vol.~14, No.~8, August~2015}%
{Shell \MakeLowercase{\textit{et al.}}: Bare Demo of IEEEtran.cls for IEEE Journals}
%



\maketitle

\begin{abstract}
	Massive captured face images are stored in the database for the identification of individuals. However,  these images can be observed { unintentionally by data examiners}, which is not at the will of individuals and may cause privacy violations.  Existing protection schemes can maintain identifiability but slightly change the facial appearance,  rendering it still susceptible to the visual perception of the original identity by data examiners.  In this paper, we propose an  effective identity hider for human vision protection, which can significantly change appearance to visually hide identity while allowing  identification for face recognizers. Concretely, the identity hider benefits from two specially designed modules: 1)  The virtual face generation module  generates a virtual face with a new  appearance by manipulating the latent space of StyleGAN2. In particular, the virtual face has a similar parsing map to the original face, supporting other  vision tasks such as head pose detection. 2) The appearance transfer module  transfers the  appearance of the virtual face  into the original face via  attribute replacement. Meanwhile, identity information  can be preserved well with the help of the disentanglement  networks. In addition, diversity and background preservation are supported to meet various requirements. Extensive  experiments demonstrate that the proposed identity hider achieves excellent performance on privacy protection and identifiability preservation.
\end{abstract}


%
\IEEEpeerreviewmaketitle

\section{Introduction}
%
%
%
%
\IEEEPARstart{F}{ace} recognition  has undergone substantial advancements in recent years. The excellent measure of safety and convenience  encourages face recognition to become a default identity management technology  in various fields, e.g., smart security and financial services.
\begin{figure}[h]
	\centering	\includegraphics[width=3.2in, keepaspectratio]{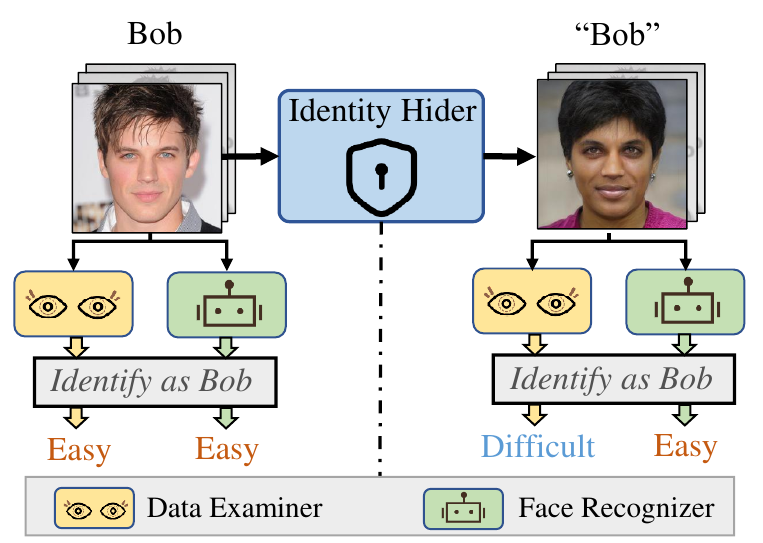}\\
	\caption{A  protection example of  the proposed identity hider. For the protected facial images of Bob, they have similar parsing maps but new appearances, rendering it  difficult for data examiners to identify them as  Bob via human vision,  but easy for face recognizers via machine vision.  }	
	\label{process}
\end{figure}


Massive  face images are captured in  databases to the facilitate  identification of individuals. Nevertheless, the  unintentional observation of these images by data examiners raises concerns about potential privacy breaches, as sensitive information could be perceived by human vision without the knowledge or consent of the individuals involved. {Although storing biometric features \cite{9499112, 9670462} can prevent the perception of human vision, these features are neither interpretable by users (vulnerable to attack)
nor compatible with iteratively updated face recognition systems.} This situation calls for the imperative development of technical solutions to protect human vision privacy while maintaining the utility of face recognition.

Plenty of  schemes \cite{yang2022study,9011585,9218980,cao2021personalized,9578149,9404267} can obfuscate the  visual content of face images to prevent  identity perception by human observers. 
However, these schemes also destroy the \textit{identifiability}  of face images for machine vision,  constraining the application of face recognition. While the  privacy-preserving face recognition scheme \cite{ji2022privacy}  allows for accurate identification, the pixelated results lack \textit{naturalness},  rendering them uninterpretable by humans and thus susceptible to potential attacks. 


Consequently, recent schemes \cite{li2021identity,yuan2022pro} were designed to modify the  visual appearance  while maintaining identifiability of faces.  Li \textit{et al.} \cite{li2021identity}  edited identity-irrelevant attributes to anonymize appearance, but the generated result loses much identity information under the adaptive mechanism, reducing the utility of identification.  To enhance the utility, PRO-Face \cite{yuan2022pro} hides the identity information of the original  face  in its obfuscated form. In this way, the protected face is visually similar to the obfuscated one from human vision, but can be still identified as the original identity by machine vision. 
However, the \textit{appearance change} of faces protected by each of these schemes \cite{li2021identity,yuan2022pro}  is slight, so that the original identity may still  be perceived by human observers. 

\begin{table*}[t]
	\centering
	\setlength{\abovecaptionskip}{-0.2cm} 
	\caption{Comparison of our scheme   and mainstream schemes . ``N/A" means ``not applicable".}
	\label{tacompare}
	\renewcommand{\arraystretch}{1}
	\setlength{\tabcolsep}{1mm}{
		\begin{tabular}{cccccccccc}		
			\\
			\toprule
			\textbf{} & Naturalness &   Identifiability&   Transferability&  Appearance Change &Real Identity& Similar Parsing Map&New background&Diversity \\		
			\midrule
			Ji \textit{et al.} \cite{ji2022privacy}&Low&High&N/A&N/A&\Checkmark&\XSolidBrush&N/A&\XSolidBrush\\
			
			Li \textit{et al.} \cite{li2021identity}&Medium&Medium&Medium&Low&\Checkmark&\Checkmark&\XSolidBrush&\XSolidBrush\\
			PRO-Face \cite{yuan2022pro}&High&High&Medium&Low&\Checkmark&\Checkmark&\XSolidBrush&\Checkmark\\
			IVFG  \cite{yuan2022generating}&High&High&N/A&High&\XSolidBrush&\XSolidBrush&\Checkmark&\Checkmark\\
			
			Ours&High&High&High&High&\Checkmark&\Checkmark&\Checkmark&\Checkmark\\

			\bottomrule
		\end{tabular}
	}
\end{table*}

Artificial intelligence-generated content (AIGC) \cite{wang2023security}  can generate virtual images that meet specified requirements, which provides a  potential  solution to human vision privacy protection. Representatively, IVFG \cite{yuan2022generating}  generates  faces with virtual identities to facilitate face recognition, which achieves satisfactory appearance changes and identifiability preservation.  Nevertheless, compared to the \textit{real identity},  the non-uniqueness of the virtual identity causes additional security issues such as difficulties in forensics.

In this paper, we aim  to significantly change the visual  appearance  while preserving identity in faces, which have two main challenges: \textit{1) Significant appearance changes:} 
Since the facial appearance  cannot be defined by computer vision,  it is not feasible  to manipulate  appearance features in a straightforward manner. Li \textit{et al.} \cite{li2021identity} tried to do it by editing  five facial attributes, which only brings about relatively weak effectiveness,  as the facial appearance covers more possible attributes, e.g.,  skin color, hairstyle, race, and age. \textit{2) Satisfactory identity preservation:}  Due to the intimate correlation between visual appearance and identity, any modification on the appearance would impact the identity information. Existing schemes \cite{li2021identity,yuan2022pro} add identity preservation constraints targeting a special face recognizer to minimize the loss of identity information. Unfortunately, they fail to support strong transferability, which means that  the identifiability would be reduced when applying an unseen face recognizer. 

To address the above challenges,  we design an effective identity hider to protect human vision privacy. 
1) For \textit{significant appearance changes}, we achieve it by modifying all face attributes instead of specific attributes.  This results in a significant change in appearance. In particular, we maintain the similar parsing map with the original face to support other  tasks, e.g., head pose recognition and glass detection. 
2) For \textit{satisfactory identity preservation}, we disentangle identity and attribute features. In this way, the loss of identity information can be minimized  when changing attributes for  appearance anonymization. Meanwhile, since the identity features are disentangled, arbitrary face recognizers can easily  extract them. 

\begin{figure}[t]
	\centering
	\includegraphics[width=3.5in, keepaspectratio]{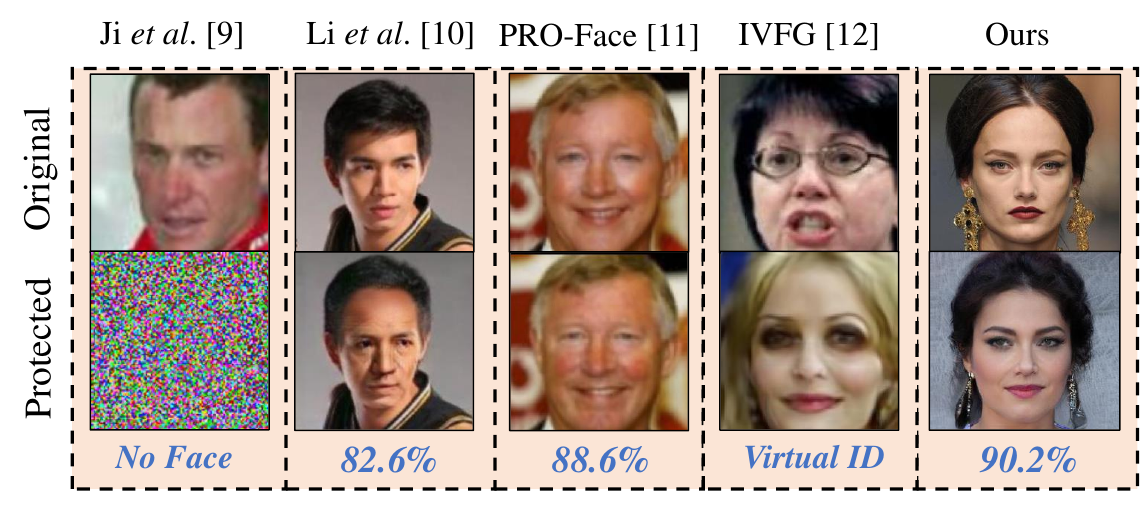}\\
	\caption{Visual samples of our hider and   mainstream schemes. Below the pair of original and protected faces  shows their identity similarities via face comparing of Face++, where the matching threshold  is  74.0\%.  }	
	\label{compare}
\end{figure}

As  illustrated in Fig. \ref{process},  the face protected by our identity hider  is visually significantly different from the original one. In this way, it is difficult for curious data examiners to identify the protected face as Bob via human vision, preventing privacy violations. Meanwhile, the face recognizer can easily determine it as Bob via machine vision, ensuring the utility of identification. In addition, our hider enables a new background, further blocking the identity perception by human vision. Table \ref{tacompare} shows the advantages of our hider over mainstream schemes, and Fig. \ref{compare} shows their corresponding visual samples.

\vspace{2 pt}

We summarize the main contributions as follows:
\begin{itemize}
	\item{We propose an  identity hider to protect human vision privacy, which can prevent human vision from the identity perception in faces while allowing  identification for machine vision.}	
	\item{The proposed identity hider can achieve satisfactory performance on  identifiability (with real identity),  transferability,  appearance change, and diversity  simultaneously, which is challenging  for existing schemes.}	
	
	\item{We design a virtual face generation module, which enables generating virtual faces with a new visual appearance and a similar parsing map. }
	\item{We design an appearance transfer module, which  can effectively transfer visual appearance to a face while preserving  identity. }

\end{itemize}

The remainder of the paper is organized as follows.
Section \ref{sec2} briefly introduces the related work about human vision privacy protection and generative adversarial networks.
{Section \ref{sec3} introduces the system model.} Section \ref{sec4} details the proposed identity hider.
Experimental results and discussions are presented in Section \ref{sec5}, and followed
by conclusion in Section \ref{sec6}.

\vspace{6 pt}
	\section{Related Work} \label{sec2}

\subsection{Human Vision Privacy Protection }
Plenty of protection schemes  \cite{yang2022study,9011585,9218980,cao2021personalized,9578149,9404267}   obfuscate the visual content of faces via pixelation, blurring, masking, or transforming.
They can prevent human vision from perceiving sensitive information, but also block content recognition by machine vision, limiting the application of computer vision. Recent schemes  enable to preserve the identifiability for machine vision, and can be divided into three groups. 

The first group of schemes 
prevent the access of all visual content of the original face image while preserving identity. { Some cryptography-based schemes can perform face recognition under secure conditions, including matrix operation-based encryption \cite{zhang2019secure}, secure multi-party computation \cite{ma2019lightweight}, and trusted execution environments \cite{zhu2020enabling}. Such schemes have a minor impact on the accuracy of recognition, but also require intolerable computational and communication overheads and special hardware deployments, which limit their practicality.  Other transformation-based schemes} focus on face recognition in the transformation domain, mainly frequency domain. Wang \textit{et al.} \cite{wang2020high} discovered that human vision relies only on low-frequency information for image recognition while machine vision relies on low-and high-frequency information. Based on it, these schemes \cite{mi2022duetface,wang2022privacy,ji2022privacy}  remove  identity-irrelevant features in the frequency domain. This ensures that the protected image is used only for face recognition, which adheres to  the principle of minimal data use. However, the unnatural results are not interpretable by humans and  do not allow for other vision tasks, e.g., face detection and attribute recognition.

To maintain naturalness, the second group of schemes \cite{li2021identity,yuan2022pro,li2021learning,10121472} only modify the facial visual  appearance while preserving identity.  Li \textit{et al.} \cite{li2021learning} learned the appearance of a reference face based on the decoupled representations, thus camouflaging the appearance of the real face. However, they did not present a specific definition of appearance, so extracting and manipulating appearance features is inherently impractical. Li \textit{et al.} \cite{li2021identity} applied class activation maps to perceive identity-irrelevant face attributes and then edit them to alter the visual appearance. Unfortunately, their experiments show that the original identity cannot be well preserved when more attributes are edited, which reduces the identifiability. PRO-Face \cite{yuan2022pro} leverages  Siamese networks to hide critical identity information of the original face into its obfuscated form. In this way,  the protected result has a similar visual appearance to the obfuscated one and the  same identity with the original one. Compared to Li \textit{et al.} \cite{li2021identity}, PRO-Face preserves more identity information and thus achieves better identification  accuracy. 

The remaining group of schemes \cite{yuan2022generating,peng2022anonym,wang2023identifiable} assign virtual identity to facilitate face recognition, which can easily enable significant  changes in visual appearance. IVFG \cite{yuan2022generating} binds  a variety of virtual faces for users in the latent space  of  StyleGAN based on keys. Such virtual faces belong to different identities from the real ones and can be used directly for privacy-preserving face recognition. Nevertheless, the irrelevant attributes like head pose  failed to be preserved, which makes it difficult to be applied in various application scenarios. Anonym-Recognizer \cite{peng2022anonym} utilizes a relationship ciphertext to reset a binary identity number for each user, and then embeds it into the appearance-anonymized face. After that, the binary identity number in the protected face is extracted for identification. Compared to IVFG, Anonym-Recognizer preserves potential attributes, improving applicability,
but requires the special recognizer, which is  incompatible with current face recognition systems. 
For this group of schemes, the adoption of virtual identity  is effective in ensuring  privacy protection  but poses other security risks, e.g., the difficulty in forensics.

\subsection{Generative Adversarial Networks}
Generative adversarial networks (GAN)  were initially proposed by Goodfellow \textit{et al.} \cite{goodfellow2020generative} and have been widely used for image generation, image editing, and other fields. A typical GAN consists of a discriminator and a generator, where the discriminator learns to distinguish between real and fake images,  and the generator learns to generate fake images to deceive the discriminator. 
Resplendently, StyleGAN \cite{karras2019style} and its upgraded versions are able to effectively decouple  high-level semantics of the generated images, enabling state-of-the-art controllable face generation.

Based on StyleGAN, one group of works \cite{shen2020interfacegan,liang2021ssflow} investigate the latent space of StyleGAN to  fine-grained manipulate the semantic attributes of the generated images.  InterfaceGAN \cite{shen2020interfacegan} identifies linear subspaces in the latent space out of specific attributes. The subspaces are later linearly manipulated to generate faces with the corresponding attributes. SSFlow \cite{liang2021ssflow}  utilizes conditional Neural Spline Flows to enhance the attribute disentanglement, which also enables the identity preservation when manipulating attributes. 

Another group of works are focused on StyleGAN inversion , which can project real images into the latent space to reconstruct the real images. I2S \cite{9008515} adopts ADAM for iterative optimization of each image to achieve high reconstruction quality but is time-consuming when inference. Instead of image-specific optimization, e4e \cite{tov2021designing}  trains an encoder to project images into the latent code. Meanwhile, it can control the trade-off between the distortion-editability and distortion-perception within the  latent space of StyleGAN.

\section{Preliminary}\label{sec3}
\begin{figure}[!b]
	\centering
	\includegraphics[width=3.5in, keepaspectratio]{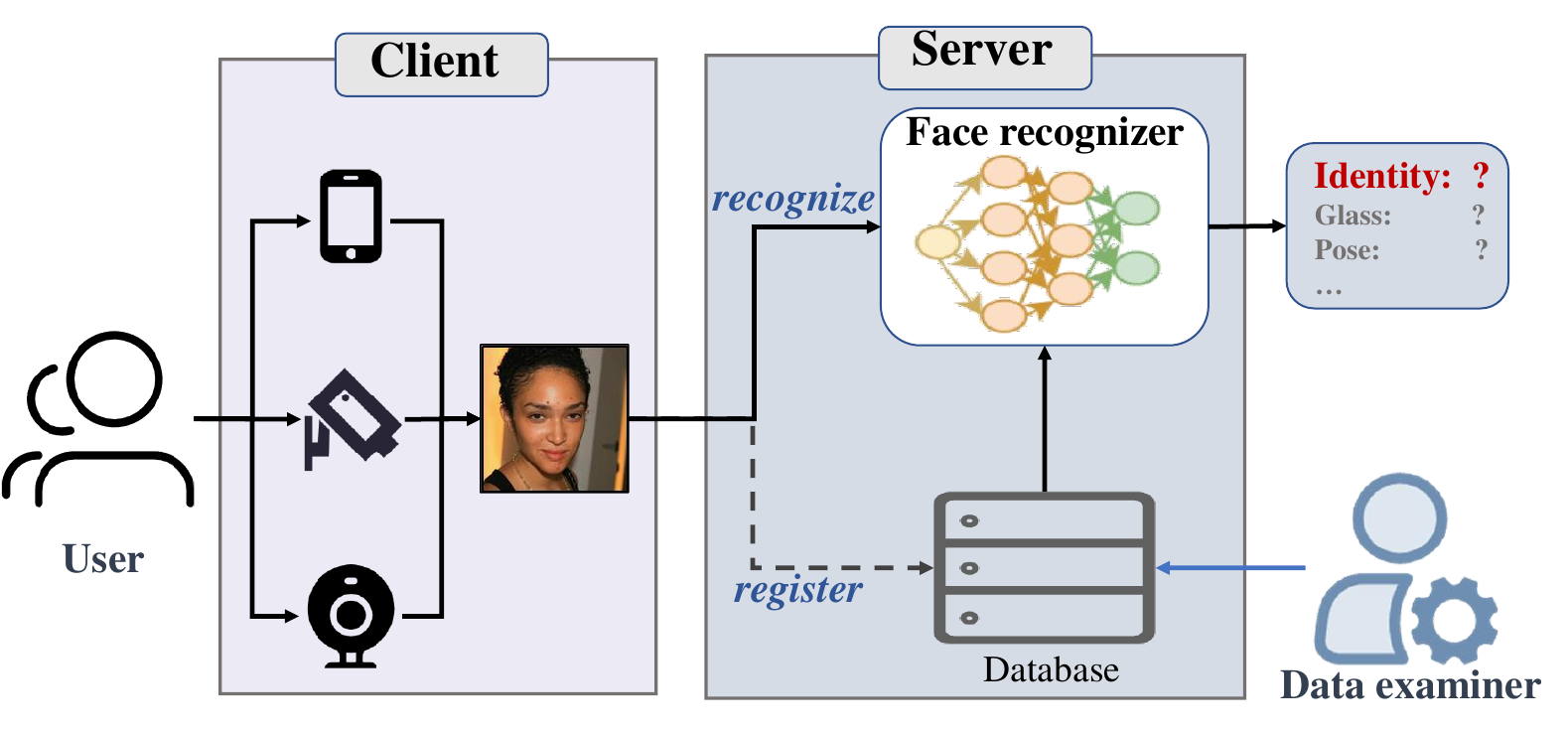}\\
	\caption{A face recognition system.}
	\label{systemmodel}
\end{figure}

\subsection{Face Recognition System}
{As illustrated in Fig. \ref{systemmodel}, a prevalent type of face recognition system currently adopts a client-server architecture. Face recognition follows the process: (1) During registration, the face of a new user is captured by the client and stored in the database.  (2) During recognition, the user's face is captured again and then recognized by comparing it with the registered face stored in the database via the face recognizer. Specifically, the system  mainly consists of four entities:
\begin{enumerate}
	\item{Client: The client is responsible for capturing face images of the user and performing preprocessing, after which the face images are transmitted to the server. }
	\item{Face recognizer: The pre-trained face recognizer is enabled by the server's powerful computing power and outputs the identity in the face image. Additionally, depending on application scenarios, the face recognizer also can recognize other attributes in the face image, e.g., pose, glass, and background.}	
	\item{Database: The database mainly stores face images during  registration. In most application scenarios, the database also records the face images during recognition to meet requirements such as forensics. }
	\item{Data examiner: The data examiner is responsible for examing the face images stored in the database to ensure data quality and standards. Since machine-based automatic methods lack robustness (especially when disturbed by adversarial and poisoning attacks), data examiners also need to sample some of the data to check it with their own naked eyes. Notably, the data examiner is assumed not to have access to the face recognizer, which can only obtain the image captured by the client as  input.}
\end{enumerate}}


\subsection{Threat Model}
{The privacy threat mainly considered   in this paper originates from the unintentional observation of the data examiner. When reviewing face images, the naked eye of the data examiner will inevitably observe the visual content of the image.  Based on the prior knowledge, the data examiner is likely to have seen the face in the image and therefore can easily match it to the person in the physical world, thus posing a privacy threat to the recognized person.  It is important to note that the data examiner, as a member of the face recognition system, does not actively initiate privacy attacks, but only unintentionally identifies people he or she has seen before when reviewing images. This unintentional recognition is also not the will of individuals.  }

{For example, let us consider a scenario in which a face image of a celebrity is captured by a face recognition system deployed at a hotel. Upon the review of this image by the data examiner, swift identification of the celebrity's identity obtains. Subsequently, the data examiner may further infer and disclose the celebrity's hotel address and fellow travelers, thus posing a significant privacy threat.}

\subsection{Design Goals}
{Therefore, the summary of our goals is designed as follows:}

{(1) It is\textbf{ easy for face recognizers }to recognize the original identity by \textbf{machine vision}. The identity features in the protected face should be satisfactorily preserved, and then the face recognizer is still able to recognize the identity with a high degree of accuracy. In addition, the impact on recognition performance for other attributes (e.g., pose, glasses, etc.) is relatively weak.}

{(2) It is \textbf{difficult for data examiners} to recognize the original identity by \textbf{human vision}. The visual appearance of the face image should be altered significantly, thereby preventing data examiners from perceiving identity directly through face appearance.}

\begin{figure*}[htb]
	\centering
	\includegraphics[width=6.5in, keepaspectratio]{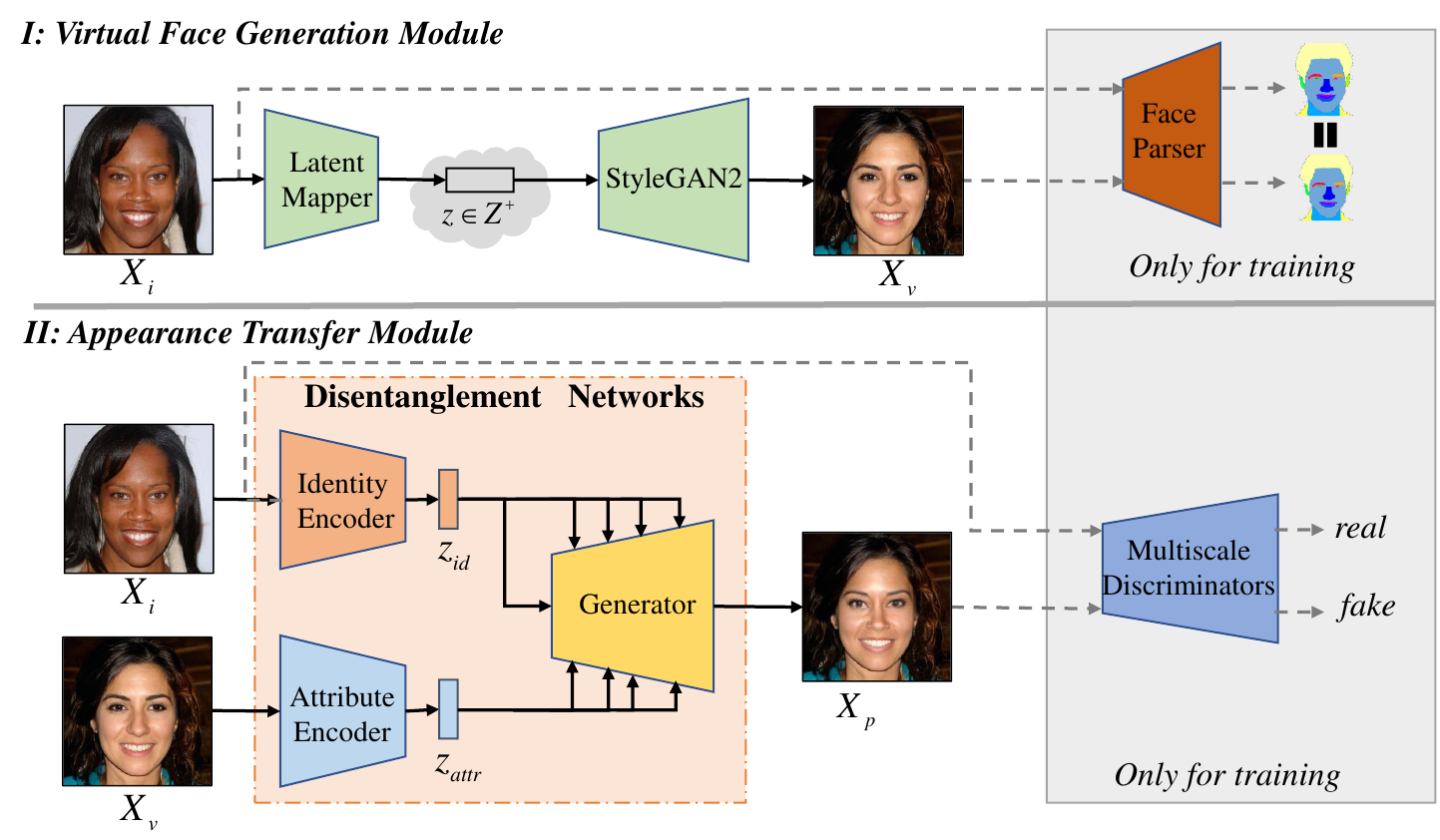}\\
	\caption{The process of the identity hider, which contains the virtual face generation module (VFGM) and the appearance transfer
		module (ATM). VFGM generates a virtual face with a new visual appearance and the similar parsing map with the original face. ATM transfers the visual appearance of the virtual face into the original face via the disentanglement networks. }
	\label{flow}
\end{figure*}
\section{The Proposed Identity Hider}\label{sec4}

Given an original face image $X_i$, the proposed identity hider can  generate a protected face image $X_p$, which has a new appearance  while preserving identity. In this way,  it  is difficult for data examiners to perceive identity via human vision,  but easy for face recognizers to extract identity via machine vision.

As illustrated in Fig. \ref{flow}, the proposed identity hider contains two modules:  virtual face generation module (VFGM) and appearance transfer module (ATM). Firstly, VFGM generates a virtual face $X_v$ with a new appearance via manipulating the latent space $\mathcal{Z+}$ of StyleGAN2. In particular, the generated face $X_v$ keeps the similar parsing map with the original face $X_i$ for supporting other low-level vision tasks such as head pose detection and glass detection.  Secondly, ATM transfers the appearance of the virtual face $X_v$  into the input face $X_i$ by attribute replacements, outputting the protected face $X_p$. Benefiting from the disentanglement networks, identity information can be  preserved  well. Moreover, additional capabilities are offered to support diverse results and preserve background,   meeting the various requirements of users.

\subsection{Virtual Face Generation Module}
In this module, we want to find a face image with  a new appearance to confuse human vision. Besides  a new appearance, we also would like  to preserve additional potential attributes to enhance the utility of protected faces. To do this, labeling some attributes for supervised learning may be a solution. However, some attributes like head pose  are complex and difficult to represent by simple classification labels. To this end, we have come up with the novel idea that we require the  parsing map to be similar with the original face, which enables us to  retain additional potential facial attributes as much as possible, particularly in maintaining the head pose.

Therefore, the goal of this module is to find a face image with  a new appearance while keeping the similar parsing map to the original face.
A naive approach is to search such faces in  face databases, which is time-consuming and may cause the privacy leakage of the real persons. StyleGAN2 \cite{karras2020analyzing} facilitates generating high-quality virtual faces based on randomly sampled vectors. Furthermore, its layered latent space $\mathcal{Z+}$ contains rich semantic information, which helps us control the generated results. Therefore, we consider generating the required virtual faces based on StyleGAN2.

Concretely, we propose a virtual face generation method built upon the $\mathcal{Z+}$ latent space of StyleGAN2. To manipulate the latent space, we design a simple latent mapper to project the input face into a latent code.  Unlike the latent encoder in GAN inversion \cite{9792208},  the designed latent mapper  only requires the latent code  to generate a new face rather than reconstruct the original face. Therefore, our latent mapper can  be more  lightweight, using only ResNet18 \cite{he2016deep} as the backbone.

To train the latent mapper, online hard example mining 
cross-entropy loss (OhemCELoss) \cite{7780458} is adopted to maintain the similar parsing map. This loss acquires samples with large cross-entropy loss to participate in training to alleviate the  problem of class imbalance. Specifically, the parsing  loss is set as follows:

\begin{equation}\mathcal{L}_{parse}=\sum_{n=1}^Ny_i^{n}log(y_v^n)+\left(1-y_i^{n}\right)log\bigl(1-y_v^{n}\bigr),
\end{equation}
where $y_i^{n}$ is the true parsing label of the original face $X_i$,  $y_v^n$ is the predicted value of virtual face $X_v$ by the face parser, and  $N$ is the number of classes.

In the practical test, the mapped latent codes would often deviate from the latent space of StyleGAN2, resulting in suboptimal generation results. For this reason, we further supplement a regularization term for constraining the distribution of latent codes to lie within the latent space. Specifically, we compute a mean latent code $\bar{z}$ by sampling 4096 latent codes, and then minimize the $L_2$ distance between the mapped  code $z$ and the mean code $\bar{z}$,
\begin{equation}
	\mathcal{L}_{reg }=\|z-\bar{z}\|_2^2.
\end{equation}

Finally, the objective for the latent mapper is formulated as follows:
\begin{equation}
	\mathcal{L}_{total}=
	\mathcal{L}_{parse}+\lambda _{reg}\mathcal{L}_{reg},
\end{equation}
where $\lambda_{reg}$ is the hyperparameter for balancing  losses.

\subsection{Appearance Transfer Module}
In this module, we expect to transfer the appearance of the virtual face into the original face. Since visual appearance cannot be defined by computer vision, manipulating appearance features is impossible. Considering that facial attributes  contain appearance information, we achieve appearance transfer by replacing all facial attributes. 

With the help of disentanglement  networks (DisenNet), we first extract the identity features $z_{id}$ of the input face $X_i$ and the attribute features $ z_{attr}$ of the virtual face $X_v$ via the identity encoder ($E_{id}$) and the attribute encoder ($E_{attr}$), respectively. Subsequently, we use the generator ($G$) to synthesize the protected face $X_p$ based on $z_{id}$ and $ z_{attr}$.  In this way, $X_p$ has a different appearance from $X_i$ while preserving the identity information.

{In the network structure design of DisenNet,  we use the pre-trained Arcface \cite{deng2019arcface} as the \textit{identity encoder}, which has a satisfactory recognition accuracy. In order to better represent different levels of semantic attributes, which contain low-level semantics like skin color and high-level semantics like gender, we design a U-Net-like network \cite{ronneberger2015u} as the \textit{attribute encoder} and represent the attribute as multi-level feature maps from the output of each layer. The \textit{generator} is stacked with multiple deconvolution layers. Nevertheless, simply concatenating the identity and attribute features as the inputs may result in ambiguous results. Thus, we utilize cascaded \textit{adaptive attentional denormalization} (ADD) \cite{li2020advancing} in the generator to couple them to adaptively adjust the effective regions of identity and attributes involvement in different  synthetic parts of the face.}

In the training process, the parameters of the identity encoder $E_{id}$ are pre-trained and thus do not require updating, while the others require. DisenNet is supervised by a weighted sum of five losses: an adversarial loss, an identity disentanglement loss, an attribute disentanglement loss, a reconstruction loss, and a visual content loss. In the following, we describe the details of each loss.

\subsubsection{Adversarial Loss}
The adoption of adversarial learning makes the generated results indistinguishable from real images, thus improving the image quality of the generated results. Considering the high resolution of the faces that we focus on, it is necessary to expand and diversify the perception range of the discriminator. Therefore, we adopt M-multiscale discriminators with hinge losses on the downsampled output images. The adversarial loss is formulated as,
\begin{equation} \mathcal{L}_{adv}=\sum_{m=1}^{M}\{ReLU(1-D_m(X_v))+ReLU(1+D_m(X_i))\},
\end{equation}
where $ReLU(\cdot)$ is the activation function, and $D_m(\cdot)$ denotes the probability output by the $m$-multiscale discriminator.

\subsubsection{Identity Disentanglement Loss} To disentangle identity, we need to enhance the similarity between the identity features of the original face $X_i$ and the generated face $X_p$,
\begin{equation}\mathcal{L}_{id}=1-cos(E_{id}(X_i),E_{id}(X_p)), \end{equation}
where $cos (\cdot, \cdot)$ represents the cosine similarity of two vectors. 

\subsubsection{Attribute Disentanglement Loss} Since attribute manipulation \cite{he2019attgan,deng2020controllable,wang2022high} aims to  modify specific attributes, cross entropy  loss is used to prompt the generated faces to have the target attributes. Unlike attribute manipulation, our work needs to replace all attributes without  the defined attribute labels. Therefore, we define the attribute features as  multi-level attribute features  obtained from the attribute  encoder. Specifically, we reduce  the $L_2$ distance between the multi-level attribute features of the protected face $X_p$ and the virtual face $X_v$ to disentangle attributes,	
\begin{equation}\mathcal{L}_{attr}=\frac{1}{2}    
	\sum_{k=1}^{K}	\left\|E_{attr}^k(X_p)-E_{attr}^k(X_v)\right\|^2_2, \end{equation}
where $E_{attr}^k(\cdot)$ is the $k$-th attribute features obtained from the $k$ layer of the attribute  encoder, and $K$ is the number of network layers in $E_{attr}$.

\subsubsection{Visual Content Loss} Unlike face swapping \cite{li2020advancing,xu2022high} which expectation is that the human observer will still perceive the swapped identity, our work expects the identity to be visually hidden.  For this, we require the visual content of the protected face $X_p$ to be similar to the virtual face $X_v$, rendering $X_p$ have a different appearance from the original face $X_i$. Concretely, we  use the pixel-level $L_2$ loss as the visual content loss to strengthen the visual similarity between $X_p$ and $X_v$,
\begin{equation}\mathcal{L}_{vs }=\frac{1}{2}\left\|X_p-X_v\right\|_2^2.
\end{equation}

\subsubsection{Reconstruction Loss} The reconstruction loss drives these decoders to learn essential facial features while removing noise and redundant information to yield a more independent representation of the features. When the input of the attribute encoder is also the original face $X_i$, we add  pixel-level reconstruction loss to train the DisenNet,
\begin{equation}\mathcal{L}_{re }=\frac{1}{2}\left\|X_i-G(E_{id}(X_i), E_{attr}(X_i))\right\|_2^2. 
\end{equation}

\subsubsection{Overall Objective} Overall, the objective for DisenNet is formulated as follows:
\begin{equation}
	\mathcal{L}_{total}=
	\mathcal{L}_{adv}+\lambda_1\mathcal{L}_{id}+\lambda_{2}\mathcal{L}_{attr}+\lambda_3\mathcal{L}_{vs}+\lambda_4\mathcal{L}_{re},
\end{equation}
where $\lambda_i$ is the hyperparameter for balancing these losses.

\subsection{Additional Capabilities}

\subsubsection{Diversity}
\begin{figure}[h]
	\centering
	\includegraphics[width=3in, keepaspectratio]{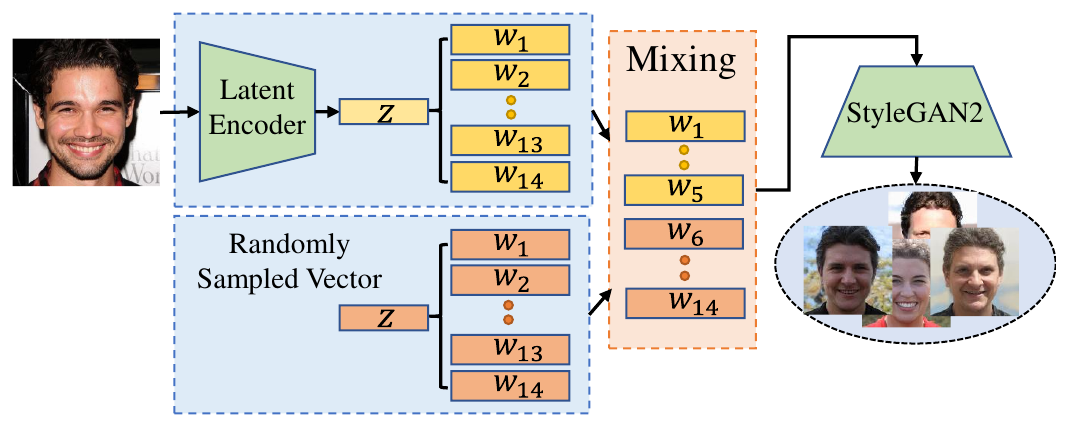}\\
	\caption{Style-mixing for diverse results. }
	\label{mix}
\end{figure}

Existing works \cite{karras2019style,9578137} have demonstrated that style-mixing can generate new faces with the similar parsing map, as the layer-wise representation of StyleGAN2 enables it to independently manipulate semantic attributes in latent space. For this, we use style-mixing to generate diverse results.  Specifically, as shown in  Fig. \ref{mix}, we calculate the latent code of the input face and randomly sample a latent code $z \in \mathcal{Z}$. Then their corresponding latent codes in $\mathcal{W+}$ are obtained. Style mixing is  performed by replacing the selected layers (6-14) of the yellow latent code  with those of the randomly generated latent codes.

\subsubsection{Background Preservation} In fact, the protected face is already identifiable for machine vision and   identity-imperceptible for human vision. Given the requirement for background consistency in special scenarios, we expect the protected face to have the same background as the original face. Therefore, we design a  simple but effective background replacement algorithm, which is illustrated in algorithm \ref{br}. We regard the area outside the head and neck as the background area. As the parsing maps of the protected image and the input image are similar,  we just separate the backgrounds of the two images directly by a  mask model $M$  which is modified by the face parser. Finally,   a simple image inpainting is used to color in the few blank areas. Of course, adopting advanced image inpainting techniques based on deep learning would be more effectively but is not necessary as only a few blank areas exist.

\begin{algorithm}[h]
	\caption{ Background Replacement.}
	\label{br}
	\begin{algorithmic}[1] 
		\REQUIRE The input face image $X_i$, the protected face image $X_p$, and the mask model $M$. \\ 
		
		\ENSURE The background-preserved protected image $X_{pb}$.\\ 
		
		\STATE $mask_{hn1}=M(X_p)$; 
		\STATE $mask_{b1}=(mask_{hn1}+1)\%2$; 

		\STATE $mask_{hn2}=M(X_i)$; 
		\STATE $mask_{b2}=(mask_{hn2}+1)\%2$; 
		
		\STATE 
		$mask=mask_{b1} \times mask_{b2}$; 
		\STATE $X_{pb}=mask \times X_b+mask_{hn1} \times X_p$;  
		\STATE $X_{pb}=inpaint(X_{pb})$; 
		\RETURN $X_{pb}$. 
	\end{algorithmic}
\end{algorithm}

\section{Experimental Results}\label{sec5}

\begin{figure*}[!ht]
	\centering
	\includegraphics[width=6in, keepaspectratio]{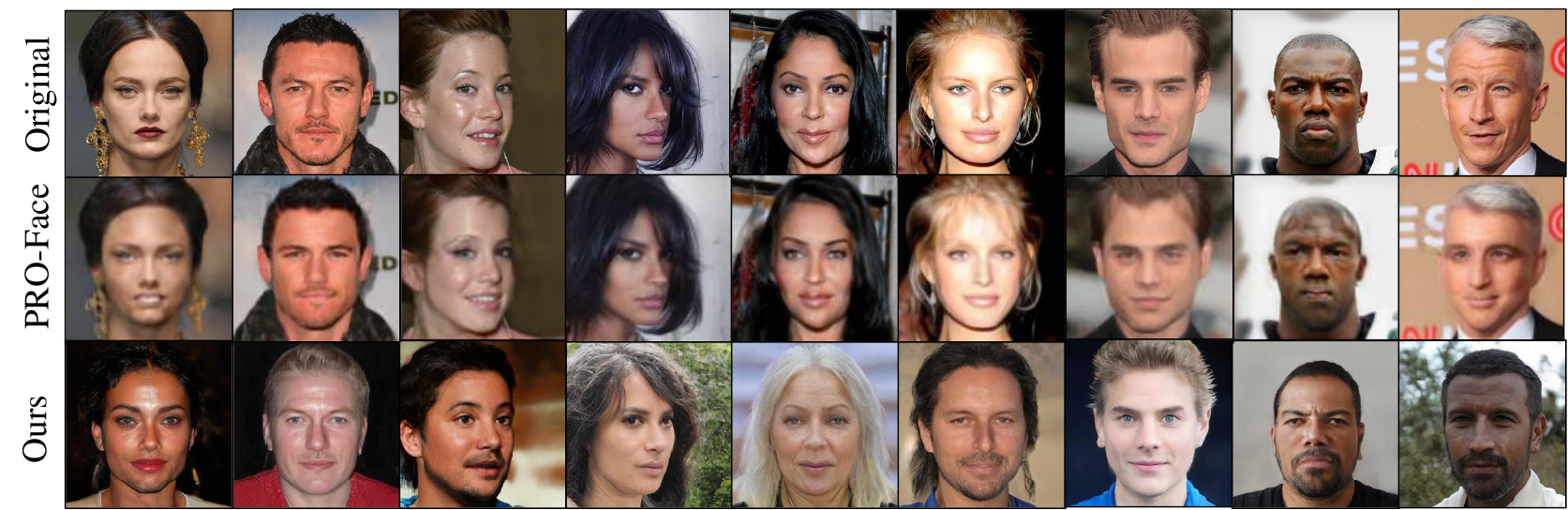}\\
	\caption{The protected samples of our identity hider and PRO-Face. Compared to PRO-Face, our hider changes the visual appearance more significantly.  Additionally, a new background further improves the ability of the identity hider to confuse human vision.
	}
	
	\label{virtual_results}
\end{figure*}

\subsection{Setup}
\subsubsection{ Dataset} 

\textbf{CelebA-HQ} is widely used for face recognition and  contains 30,000 aligned samples at 1024$\times$1024 solution. We  revised each image to 256$\times$256 and used 27,000 images sorted by indices as training data, 1,000 images as validation data, and the remained images as testing data for both the latent mapper and DisenNet. \textbf{VGGFace2} is downloaded from Google and contains over 3.31 million images of 9,131 identities.  We used MTCNN to detect faces and then revised the images to 256$\times$256. This dataset has been divided into the training and testing data, which are only used for the parameter optimization of DisenNet. In the identifiability preservation evaluation, we randomly selected 2,000 face pairs with the same identities and 2,000 face pairs with different identities in CelebA-HQ.

\subsubsection{ Implementation Details}  The latent mapper is trained by Adam optimizer with $\beta_1=0$ and $\beta_2=0.99$. The batch size is set to 16 , the initial learning rate is set to 0.0001 and the weighting hyperparameter is set to $\lambda_{reg}=30$. DisenNet is also trained by Adam optimizer with $\beta_1=0$ and $\beta_2=0.99$. The batch size is set to 8, the initial learning rate is set to 0.0004 and the weighting hyperparameters are set to $\lambda_{1}=10$, $\lambda_{2}=20$, $\lambda_3=10$, and $\lambda_4=10$. We perform one generator update after one discriminator update. 

\subsubsection{ Baseline}  We consider the state-of-the-art framework PRO-Face \cite{yuan2022pro} as our baseline. To make the experiments comparable, we chose the image obfuscation as FaceShifter and the face recognizer as IResNet100. In addition, when generating obfuscated faces, we choose a random target face instead of using the same target face. 

\subsection{Evaluation on  Protection for Human Vision Privacy}

\subsubsection{Quantitative Analysis} We use LPIPS, SSIM, MAE, and RMSE  to measure the similarity between the protected images and original images. Higher LPIPS (lower SSIM, higher MAE, or higher RMSE) represents
a low similarity, which indicates a stronger ability of privacy protection.

\begin{table}[!t]
	\begin{center}
		\caption{Similarity measures between  original images and protected images by our hider and PRO-Face. }
		\label{tab1}
		\begin{tabular}{ccccc}
			\toprule
			
			&LPIPS$\uparrow$&SSIM$\downarrow$&MAE$\uparrow$&RMSE$\uparrow$\\
			\midrule
		{	PRO-Face (CelebA-HQ)}   & {0.196} &{0.828}&{0.094}&{0.137}\\
			
			Ours (CelebA-HQ)  &\textbf{0.559}&\textbf{0.306}&\textbf{0.256}&\textbf{0.323}\\
			\midrule
		{	PRO-Face (VGGFace2) }  &{ 0.159} &{0.821}&{0.092}&{0.134}\\
			
			Ours (VGGFace2)  &\textbf{0.588}&\textbf{0.315}&\textbf{0.251}&\textbf{0.316}\\
			
			\bottomrule
		\end{tabular}
	\end{center}
\end{table}

The results are shown in Table \ref{tab1}, where the LPIPS and SSIM of the PRO-Face are from the original paper. It is clear that our results outperform PRO-Face on all metrics. Specifically, our hider is nearly five times that of PRO-Face in LPIPS, MAE, and RMSE, and nearly half that of PRO-Face in SSIM. Therefore, our hider can significantly change the visual content of the image, and the facial appearance changes accordingly.  Since PRO-Face only modifies fewer facial areas, the appearance change is slight, and thus the similarity is high. Our hider can change the entire head area and background, which effectively prevents human vision from perceiving identity.

\subsubsection{Qualitative Analysis}

Firstly, we show the protected images generated by our hider and PRO-Face in Fig. \ref{virtual_results}, where the resolution of PRO-Face is 112$\times$112 and our resolution is 256$\times$256. As can be seen, PRO-Face only  slightly modifies the facial appearance. Thus, it is easy to perceive that the identity of the protected face is consistent with the original one, when a human observer compares them. Our results nearly generate  new heads, which makes it  difficult for human vision to perceive the real identities. Unlike 	Li \textit{et al.} \cite{li2021identity} which only changes five attributes, we enable modifications on more attributes, including   skin color, beard, age, race, gender, hairstyle, and makeup.  Additionally, the identity hider generates a new background, further improving the ability of confusing human vision. When the background is required to be preserved, a feasible result is shown in \ref{secback}. In general, our identity hider achieves a significant change in appearance, which can effectively confuse human vision.


Secondly, we conduct a user study to evaluate whether the identity hider could confound real persons. We randomly selected 30 identities, each of which was constructed as a user recognition test.  A user recognition test presents a probe image (the protected face) along with eight candidate images and a ``No Match" option. The eight candidates include an image with the same identity as the probe and  other images from other identities. 
Twenty-six participants were invited to select  the one that best matches the probe's identity from candidate images or select ``No Match". Table \ref{tab2} shows the results of the user study, which  indicates that the protected images can reduce the recognition accuracy of human vision. Therefore, the proposed identity hider can effectively protect human vision privacy.

\begin{table}[!t]
	\begin{center}
		\caption{Experimental results of the user study on human recognition.  }
		\label{tab2}
		\begin{tabular}{cll}
			\toprule
			
			&Accuracy$\downarrow$ &No Match$\uparrow$\\
			\midrule
			Original & 0.915&0.051\\
			
			Identity hider  & 0.254 (-0.661)&0.554 (+0.503)\\
			\bottomrule
		\end{tabular}
	\end{center}
\end{table}

\begin{figure*}[t]
	\centering
	\includegraphics[width=6in, keepaspectratio]{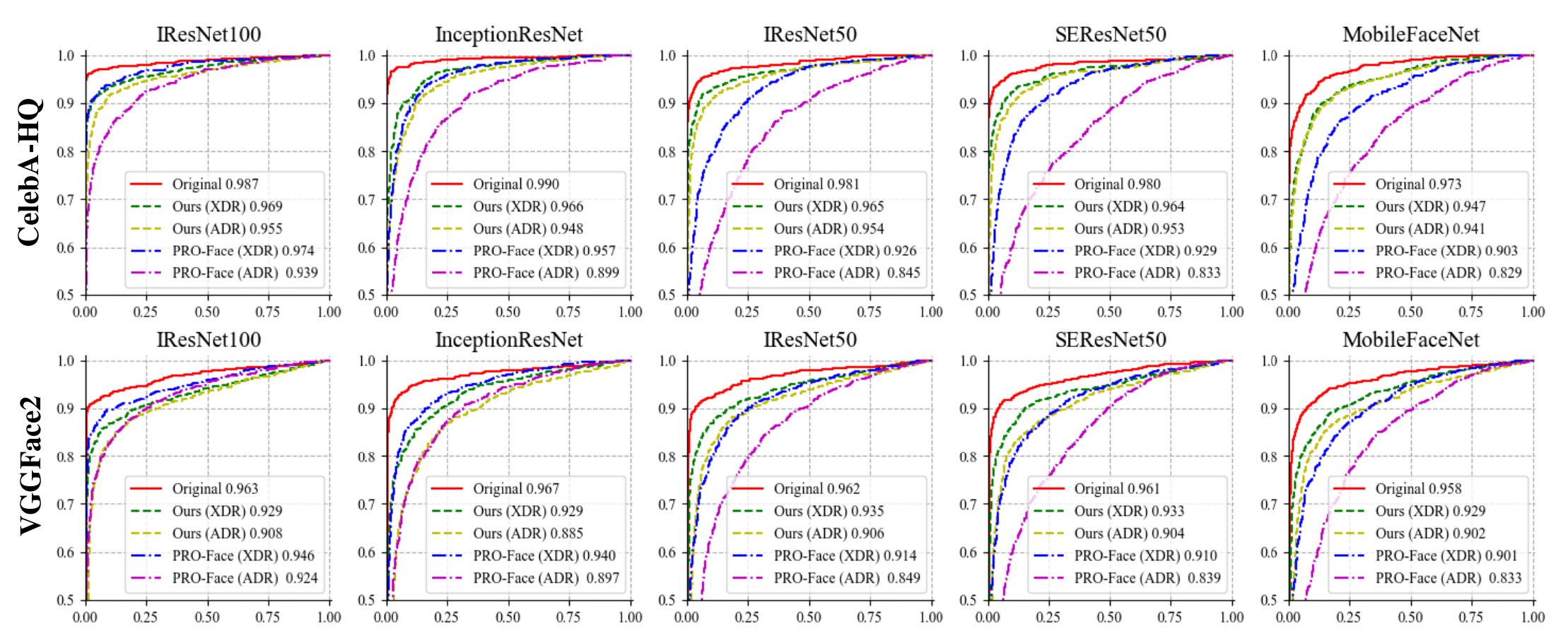}\\
	\caption{ROC curves  of the five face recognition models on CelebA-HQ and VGGFace2 under different
		image domains: original-domain, anonymized-domain
		(ADR), cross-domain (XDR).}
	\label{identity}
\end{figure*}


\subsection{Evaluation on Identifiability Preservation} 
To evaluate the identifiability, we conducted face verification experiments on various face recognition models, including MobileFaceNet \cite{chen2018mobilefacenets}, InceptionResNet \cite{szegedy2017inception}, IResNet50 \cite{duta2021improved}, SEResNet50 \cite{8578843}, and IResNet100 \cite{duta2021improved}, all of which have satisfactory recognition accuracy. Referring to the work of PRO-Face, we also considered two different recognition scenarios: 1) Anonymized-domain recognition (\textbf{ADR}): In the pair of images  for face verification, both of them are protected by our identity hider or PRO-Face; 2) Cross-domain recognition (\textbf{XDR}): In the pair of images  for face verification, only one of them is protected by our identity hider or PRO-Face.

Fig. \ref{identity} shows the receiver operating characteristic (ROC) curves of the face verification. According to the results,  PRO-Face outperforms our identity hider on IResNet100 (the model used in training) and InceptionResNet, but is inferior to ours on other models. Particularly  under ADR, the ROC value of PRO-Face on other models is lower than 0.85, which indicates that it has a low identifiability.  This is mainly because it adds identity preservation constraints to one face recognizer in training, which is difficult to generalize to other recognizers. In contrast, our identity hider utilizes disentanglement learning, which promotes strong transferability. As Fig. \ref{identity} shows, our identity hider obtains  ROC values of higher than 0.9 on two datasets, five face recognition models, and two recognition scenarios. Therefore, the identity hider preserves satisfactory identifiability and has stronger transferability than PRO-Face.

\begin{figure}[t]
	\centering
	\includegraphics[width=2.2in, keepaspectratio]{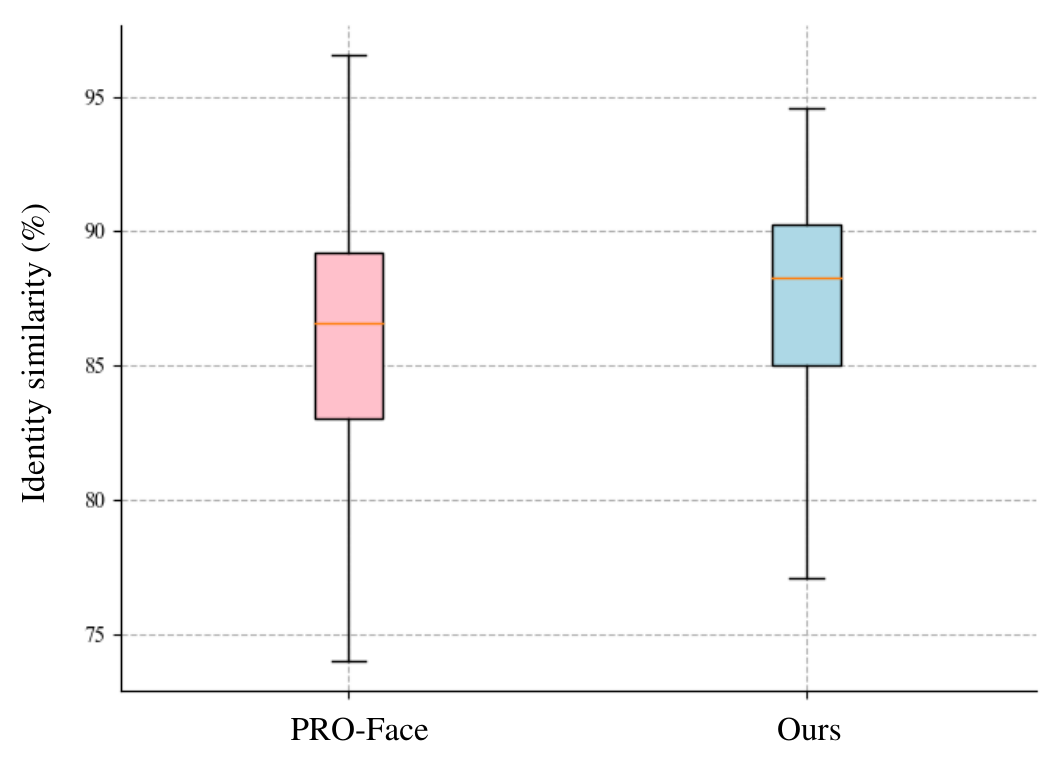}\\
	\caption{Identity similarities between original faces and protected faces via Face++, where the matching threshold  is  74.0\%.}
	\label{box_plots}
\end{figure}

In addition, we measured the identity similarity between the original face and the protected face on  Face++\footnote{https://www.faceplusplus.com.cn/face-comparing/}, where they were recognized as the same identity when the similarity was greater than the matching threshold (74.0\%). To reveal the distribution properties of the similarity scores, the corresponding box-plots are shown in Fig. \ref{box_plots}. It is clear that, compared with PRO-Face, our identity hider exhibits smaller score deviations while obtaining higher  average scores. Moreover, our results all exceed  74.0\%, while the minimum value of PRO-Face is close to the threshold. Therefore, our identity hider can preserve a higher degree of identifiability.

\subsection{Evaluation on Preservation of Similar Parsing Maps } 
The protected faces generated by  Yuan \textit{et al.} \cite{yuan2022generating}  have completely different parsing maps from the original ones, making them unable to support other vision tasks such as head pose detection. Our identity hider can preserve similar face semantic maps, further enhancing the utility of protected faces. Due to the lesser change in appearance, PRO-Face is able to retain the parsing map better compared to our hider.  Therefore, we do not require a comparison with PRO-Face but only take it as a reference. If our performance is not significantly different from PRO-Face, we can state that the similar parsing map can be preserved by our identity hider. 
\begin{figure}[!ht]
	\centering
	\includegraphics[width=3.5in, keepaspectratio]{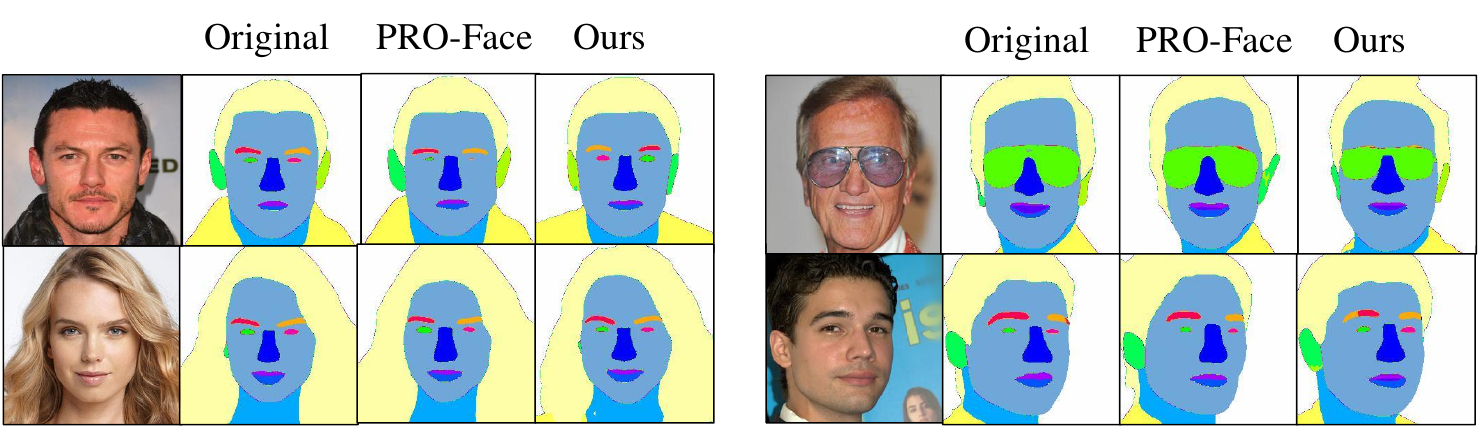}\\
	\caption{The parsing maps of the protected faces. }
	\label{visul_parsing}
\end{figure}

\begin{figure}[!h]
	\centering
	\includegraphics[width=3.5in, keepaspectratio]{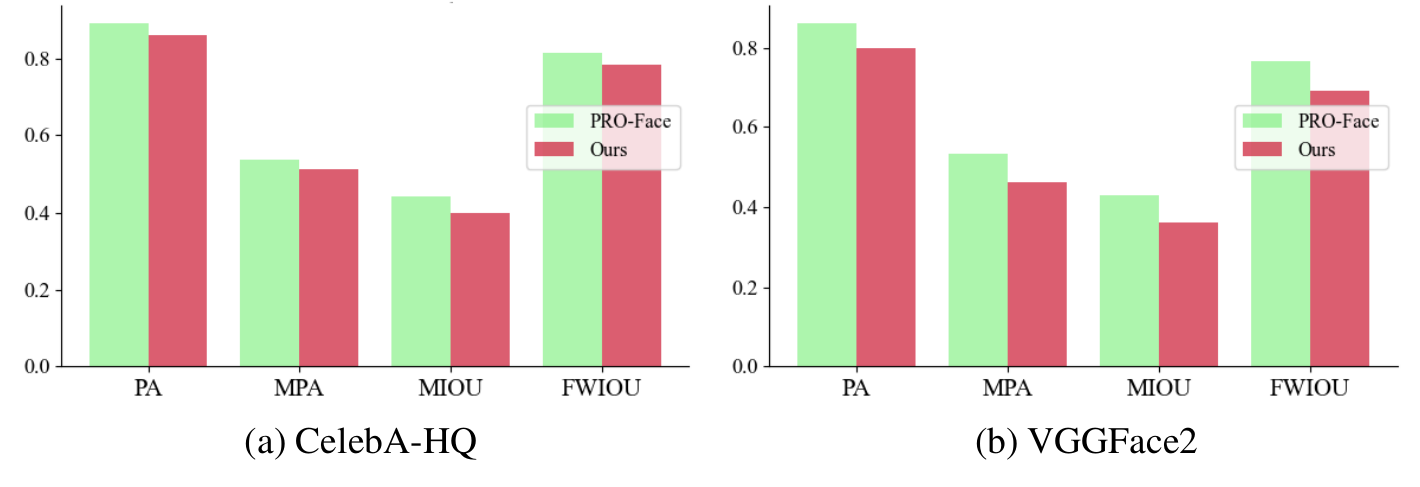}\\
	\caption{
		The results of multiple measures to evaluate the  similarity of parsing maps, including PA, MPA, MIOU, and FWIOU. }
	\label{parsing}
\end{figure}

Fig. \ref{visul_parsing} shows several parsing maps of the protected faces. It can be intuited that the parsing maps in each column are similar to each other. For further quantitative analysis, we selected {four important semantic segmentation metrics \cite{lateef2019survey},} including pixel accuracy (PA), mean pixel accuracy (MPA), mean intersection over union (MIoU), and frequency weighted intersection over union (FWIOU). In particular, the regions of the parsing map are divided into 19 classes, containing various small regions. Fig. \ref{parsing} shows the results of  evaluating similar parsing maps between the original faces and the protected faces generated by our hider and PRO-Face. As  can be seen,  our results do not significantly differ from PRO-Face. Meanwhile, on both metrics PA and FWIOU, our results are greater than 0.8. Since the 19 maps contain small facial regions, e.g., nose and  eyes,  our hider and PRO-Face  both perform unsatisfactorily on the average metrics (MPA and MIOU). Overall, the parsing maps of our protected result can not  maintain consistency with the original face but hold  a strong similarity.

In addition, we explore the preservation of potential attributes with the help of Face++'s face detection API\footnote{https://api-cn.faceplusplus.com/facepp/v3/detect}. Specifically, we detected the preservation of head pose (pitch, roll, and yaw), glass, and emotion. {As shown in Table \ref{tabdetect}, both ours and PRO-face }are capable of preserving a certain degree of head pose, wherein the angular offset in each direction does not exceed $4^\circ$. In particular, the glasses can be accurately detected. Unfortunately, facial emotions cannot be retained with high accuracy by our hider and PRO-Face. In fact, facial emotions may  not have a strong correlation with perceiving  identity by human vision, and thus can be further retained with an  exploration. More discussions can be found in \ref{utilit}. Overall, our hider is able to preserve some attributes while there is still room for optimization.

		
		

\begin{table}[t]
	\begin{center}
		\caption{Preservation of potential attributes, including head pose, glass, and emotion.}
		\label{tabdetect}
		\begin{tabular}{cccccccc}
			\toprule
			&\multicolumn{3}{c}{Head Pose ($^\circ$)}&\multirow{2}{*}{Glass}&\multirow{2}{*}{Emotion}\\
			&Pitch&Roll&Yaw\\
			\midrule
			PRO-Face (CelebA-HQ)&2.612&1.776&2.712&1.000&0.789\\
			Ours (CelebA-HQ)&3.592&2.315&3.799&1.000&0.730\\
			\midrule
			PRO-Face (VGGFace2) &2.772&1.921&3.135&1.000&0.762\\
			
			Ours (VGGFace2) &3.519&2.393&3.831&1.000&0.692\\
			
			\bottomrule
		\end{tabular}
	\end{center}
\end{table}

\subsection{Evaluation on Additional Capabilities} \label{secback}
\subsubsection{Diversity}

\begin{figure}[!ht]
	\centering
	\includegraphics[width=3.2in, keepaspectratio]{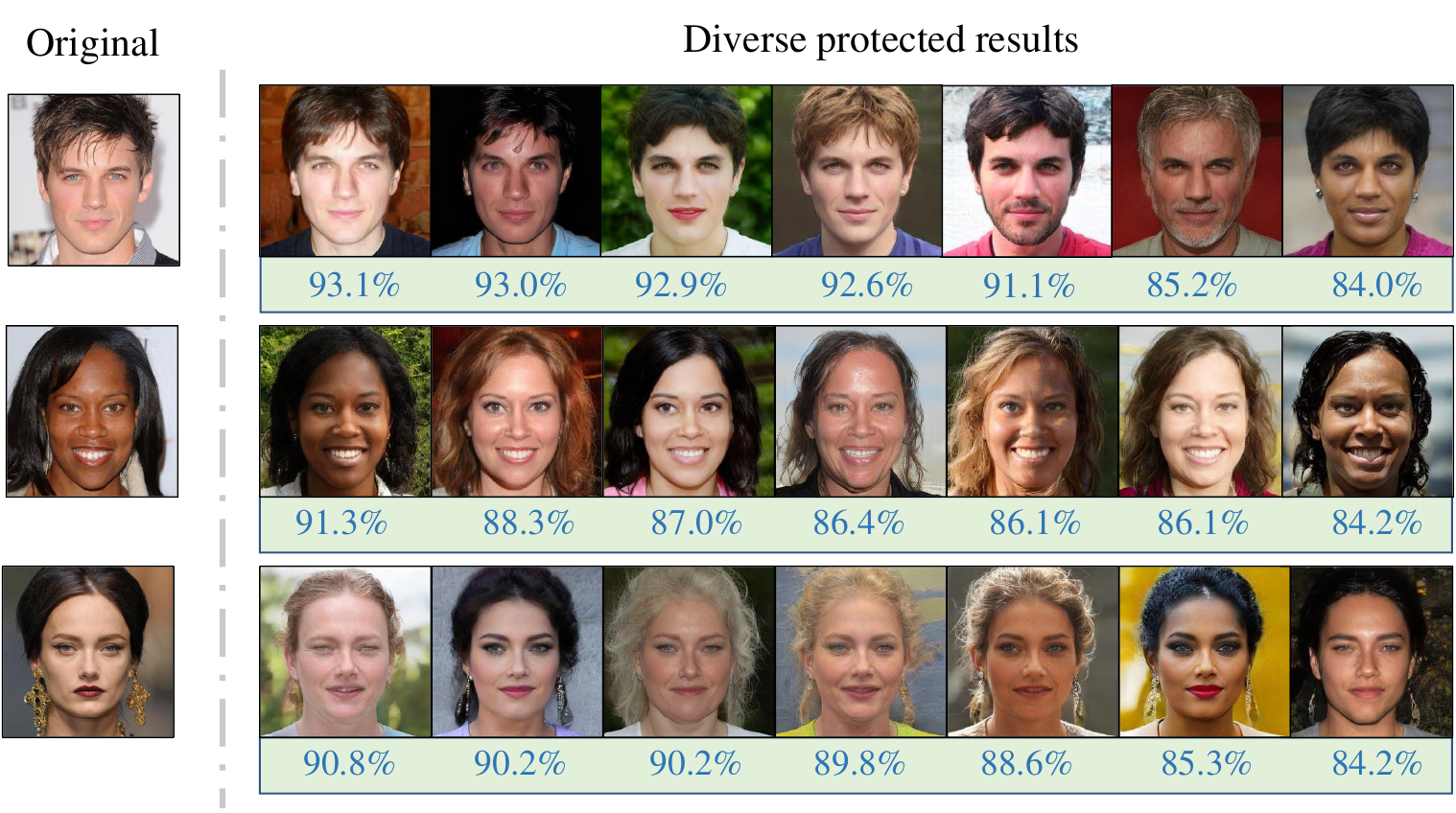}\\
	\caption{Diverse protected results. Below each protected face is the identity similarity by Face++'s face comparing, (matching threshold  is  74.0\%) }
	\label{diversity}
\end{figure}

In Fig. \ref{diversity}, we show diverse faces protected by our identity hider.  In particular, we measured the verification results between each protected face and the original face via Face++.  When the result is greater than 74\%, the two faces can be considered to belong to the same identity.  It can be noticed that our verification results are much more than 74\%, which can indicate that our diverse results can retain identifiability.  Moreover, each of these protected
results has a new appearance that effectively prevents human visual vision from identity perception. Intuitively, they also maintain similar parsing maps with the original faces, which benefits from controllable latent space  $\mathcal{W+}$ of StyleGAN2.

\subsubsection{Background Preservation}

\begin{figure}[!t]
	\centering
	\includegraphics[width=3.5in, keepaspectratio]{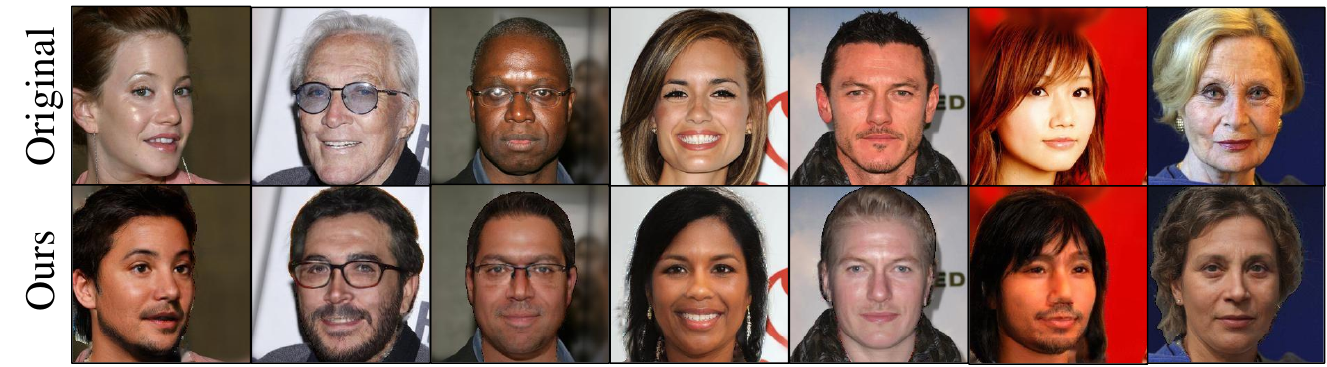}\\
	\caption{The    background-preserved results of our identity hider.
	}
	\label{virtual_back}
\end{figure}

Since the protected face has a similar parsing map to the original face, we only use a simple but effective  algorithm to preserve the original background. Fig. \ref{virtual_back} illustrates the background-preserved results. We can observe the protected face can maintain the original background. Although it can be found a slight border between the face region and the background through careful observation, this does not affect the overall perception of the user experience. In particular, when the background is preserved, it seems easier for human vision to perceive that the original image and the protected image come from the same identity. This  emphasizes the need for our identity hider to generate a new background,  which can strengthen the blocking of identity perception by human observers.


\vspace{-6 pt}
\subsection{Additional Experiments}

\subsubsection{Trade-off between Privacy and Utility}

\begin{figure}[!ht]
	\centering
	\includegraphics[width=3.5in, keepaspectratio]{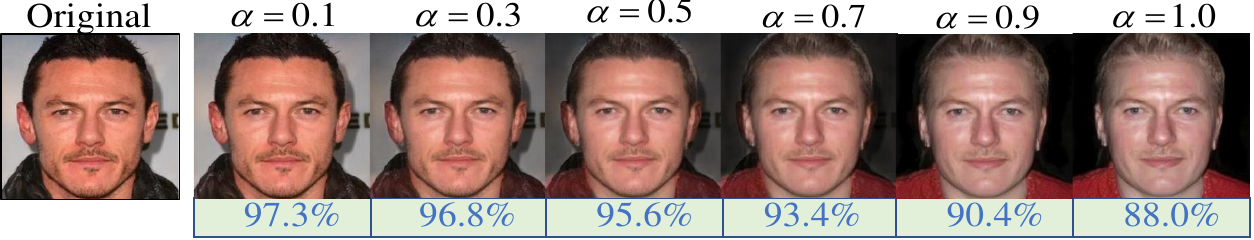}\\
	\caption{Trade-off between privacy and utility.}
	\label{trade-off}
\end{figure}

We balance the privacy (human vision) and utility (identifiability) by controlling the ratio of changed attributes of the protected face. Specifically, in the appearance transfer module, we set the  attributes used for replacement to be a combination of the attributes of the original face and the virtual face. The attributes of the virtual face account for $\alpha$ and the attributes of the original face account for $1-\alpha$. We set the $\alpha$ from 0.1 to 1 at intervals of 0.2 to generate the corresponding protected face. The results are shown in Fig. \ref{trade-off}. As $\alpha$ increases, the appearance of the protected face gradually changes while the identifiability slowly decreases. Notably,  the face verification results for all protected faces via Face++ are  above the threshold (74.0\%), benefiting from the identity disentanglement.

\subsubsection{Ablation Experiment}  \label{abla_ex}

Compared to typical face swapping, the designed DisenNet has an additional visual content loss. This loss can make the appearance of the protected face deviate from the appearance of the original face, thus obfuscating human vision. Ablation experiments were carried out to verify the effectiveness of this loss. Specifically, we remove the visual content loss (``w/o vs'') to generate the protected results, which are shown in Fig. \ref{trade-off}.  It can be seen, ``w/o vs'' improves identity similarity while enabling a new appearance, which is consistent with the goal of face swapping. Although the  difference in SSIM between  ``w/o vs'' and ours is small,  the perceived identity of ``w/o vs'' is more similar to original identity than the perceived identity of ours by human vision. In other words, our appearance changes more noticeably, including changes in skin color, beard.  Even though identity similarity of ours decreased a little, it is still much greater than the matching threshold (74\%).  Therefore,  the visual content loss can  enhance the ability of our identity hider to obfuscate human vision.

\begin{figure}[!ht]
	\centering
	\includegraphics[width=3.2in, keepaspectratio]{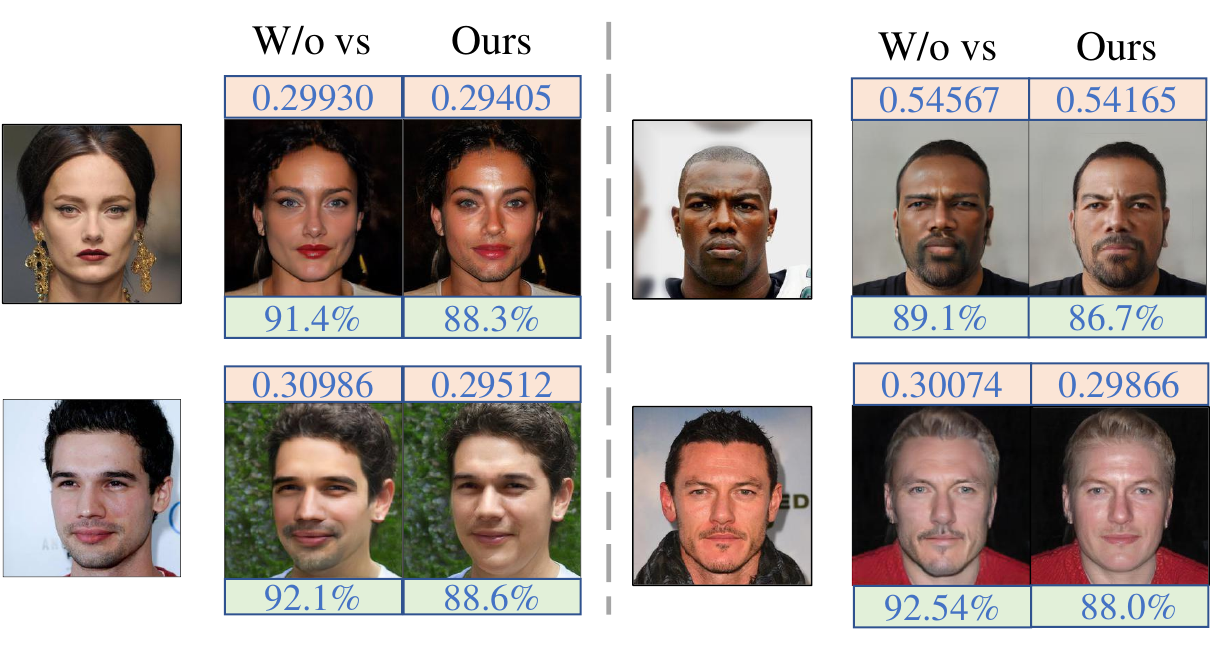}\\
	\caption{Ablation experiments. Above each protected face is the value of SSIM, and below each protected face is the identity similarity.}
	\label{abalation}
\end{figure}

\subsection{Discussion} \label{discuss}

\subsubsection{ Evaluating Human Vision Protection}  
Identity perception via human vision is a complex perceptual process that involves an integrated neural reasoning process. Although we have significantly altered the appearance of the protected face, some participants in the user study were still able to accurately perceive the original identity. Since the participants were diverse (different priori knowledge or reasoning abilities) and limited in number, it was difficult to collect accurate results to assess our hider. Thus, quantitative evaluation is more important.

However, there are no quantitative metrics for evaluating the protection of human vision  privacy. In our paper, we can only adopt the  metrics that is used to evaluate the similarity between images, which can not accurately reflect our protection effectiveness. In the \ref{abla_ex}, although the SSIM values are close before and after removing visual content loss, the appearance changes are different. Therefore, it is necessary for researchers to study the metrics for human vision protection.
\subsubsection{Utility Enhancement} \label{utilit}
Our identity hider focuses on preserving the utility of identifiability. In addition, we maintain similar parsing maps as much as possible to preserve head poses and glasses, promoting utility enhancement. Unfortunately, it fails to have a high accuracy in emotion recognition, which  has no strong correlation with appearance. In other words, perceiving identity by human vision is not highly affected by emotion, so emotion can be preserved.  However facial appearance (or the perception ability of human vision) cannot be defined by machine vision, making it challenging to list potential irrelevant attributes to be preserved. Hence, subsequent research could explore the definition of appearance to further enhance the utility of protected faces.

\subsubsection{Adaptive Attacker}
{ In addition to unintentional visual observations, the malicious data examiner may use off-the-shelf identity extractors to initiate active attacks. In this way, the attacker can  extract the correct identity feature and match the registered face in the database. However, since the registered face image is protected (significant visual appearance changes), the attacker cannot  find a matched user in the physical world based on their prior knowledge. Therefore, the personal privacy is protected.}
 
{Furthermore, we assume that the attacker collects a large number of unprotected face images from the physical world. In this way, the attacker is able to locate the user in the physical world by matching the protected face to the unprotected face in the physical world using the identity extractor. However, due to the obvious visual difference between the protected face and the user in the physical world, it is difficult for the attacker to believe that the protected face is from the physical user. The attacker thinks  that ``seeing is believing" and therefore is more confident that the identity extractor is broken. Therefore, the personal privacy still can be protected.}

\section{Conclusion}\label{sec6}

In this paper,  we present an identity hider to protect human vision privacy. It enable to prevent the  identity perception by data examiners while preserving high identifiablity for face recognizers.  Specifically, the identity hider contains two modules, which can generate virtual faces and transfer appearance, respectively. Based on the above modules, the identity hider is able to significantly change the visual appearance to obfuscate human vision while preserving identity for human vision. In addition, diversity and background preservation are supported to meet the different requirements. We also conducted excessive  experiments to demonstrate  the effectiveness of the proposed identity hider. 

\small
\bibliographystyle{IEEEtran}
\bibliography{reference}


%




\ifCLASSOPTIONcaptionsoff
  \newpage
\fi



%

%

\begin{IEEEbiography}[{\includegraphics[width=1in,height=1.25in,clip,keepaspectratio]{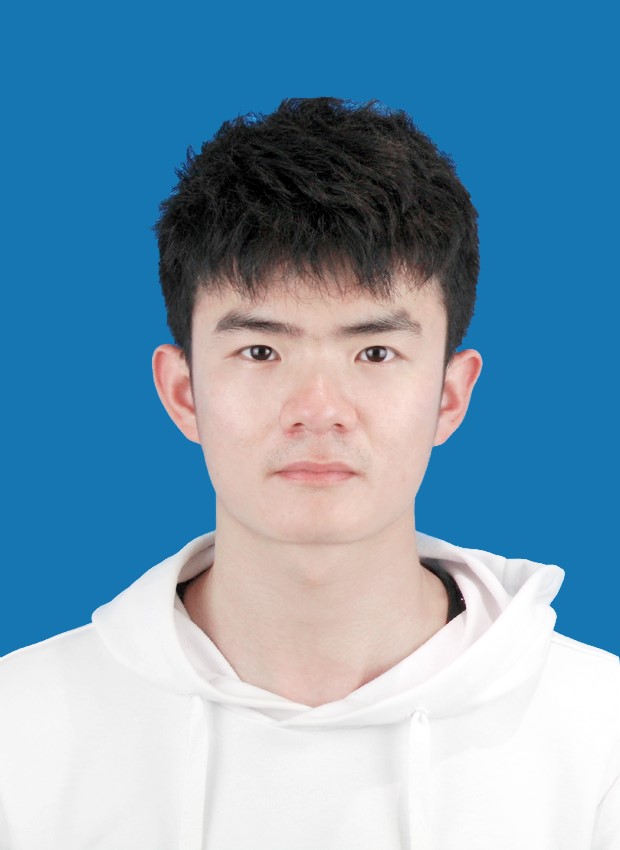}}]{Tao Wang} received the M.S. degree in  cyberspace security from the College of Computer Science and Technology, Nanjing University of Aeronautics and Astronautics, Nanjing, China, in Apr. 2022. He is currently working toward the Ph.D. degree in cyberspace security with the College of Computer Science and Technology, Nanjing University of Aeronautics and Astronautics, Nanjing, China. His current research interest is image privacy.
\end{IEEEbiography}

\begin{IEEEbiography}[{\includegraphics[width=1in,height=1.25in,clip,keepaspectratio]{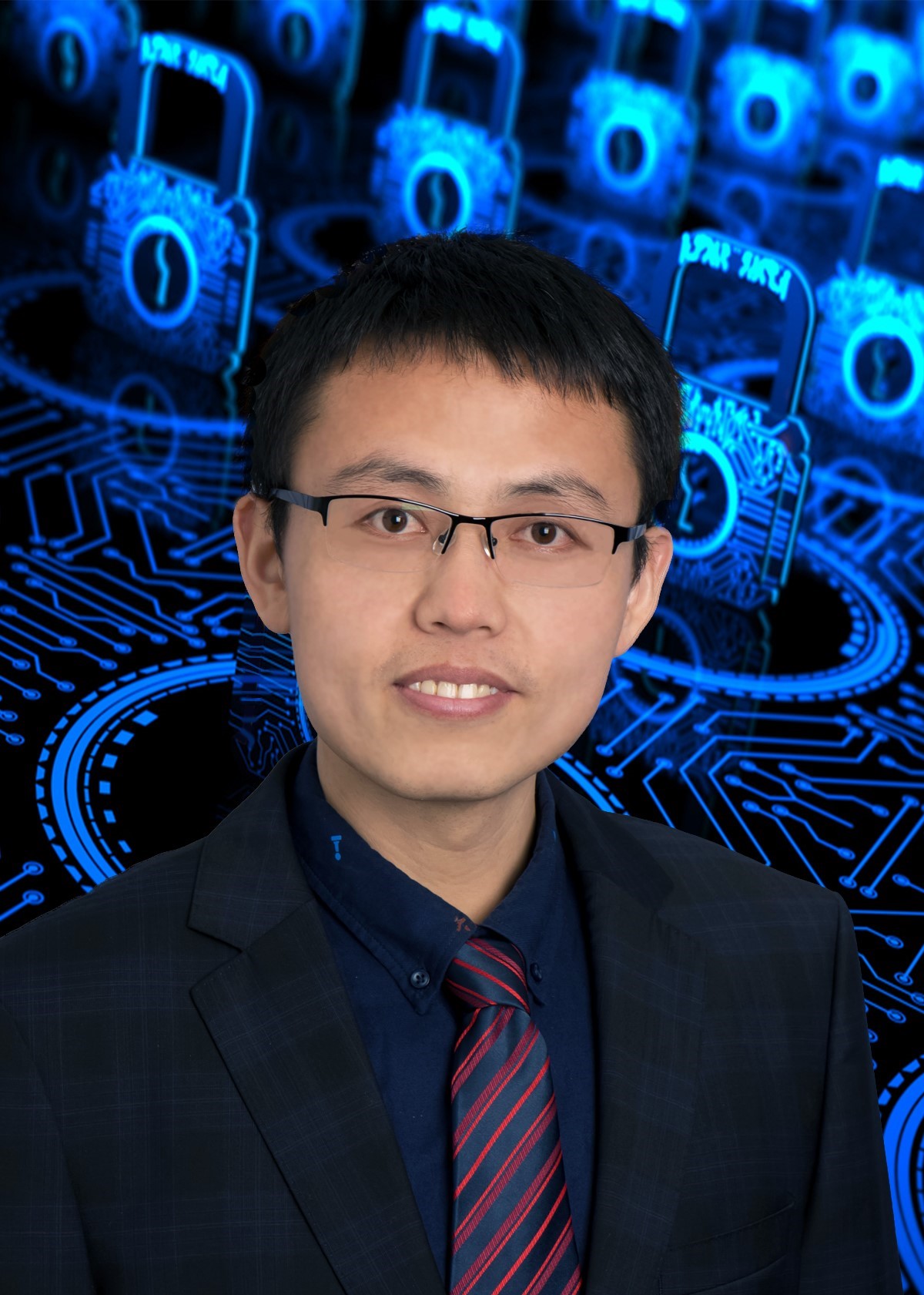}}]{Yushu Zhang} received the Ph.D. degree from the College of Computer Science, Chongqing University, Chongqing, China, in December 2014. He is currently a Professor with the College of Computer Science and Technology, Nanjing University of Aeronautics and Astronautics, Nanjing, China. His research interests include multimedia security, artificial intelligence security, and blockchain.\end{IEEEbiography}

\begin{IEEEbiography}[{\includegraphics[width=1in,height=1.25in,clip,keepaspectratio]{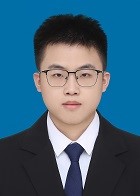}}]{Zixuan Yang} received the B.E. degree from the College of Computer Science and Technology, Nanjing University of Aeronautics and Astronautics, Nanjing, China, in June 2021, where he  is currently pursuing the M.S. degree. His current research interests include multimedia security, adversarial attack.
\end{IEEEbiography}

\begin{IEEEbiography}[{\includegraphics[width=1in,height=1.25in,clip,keepaspectratio]{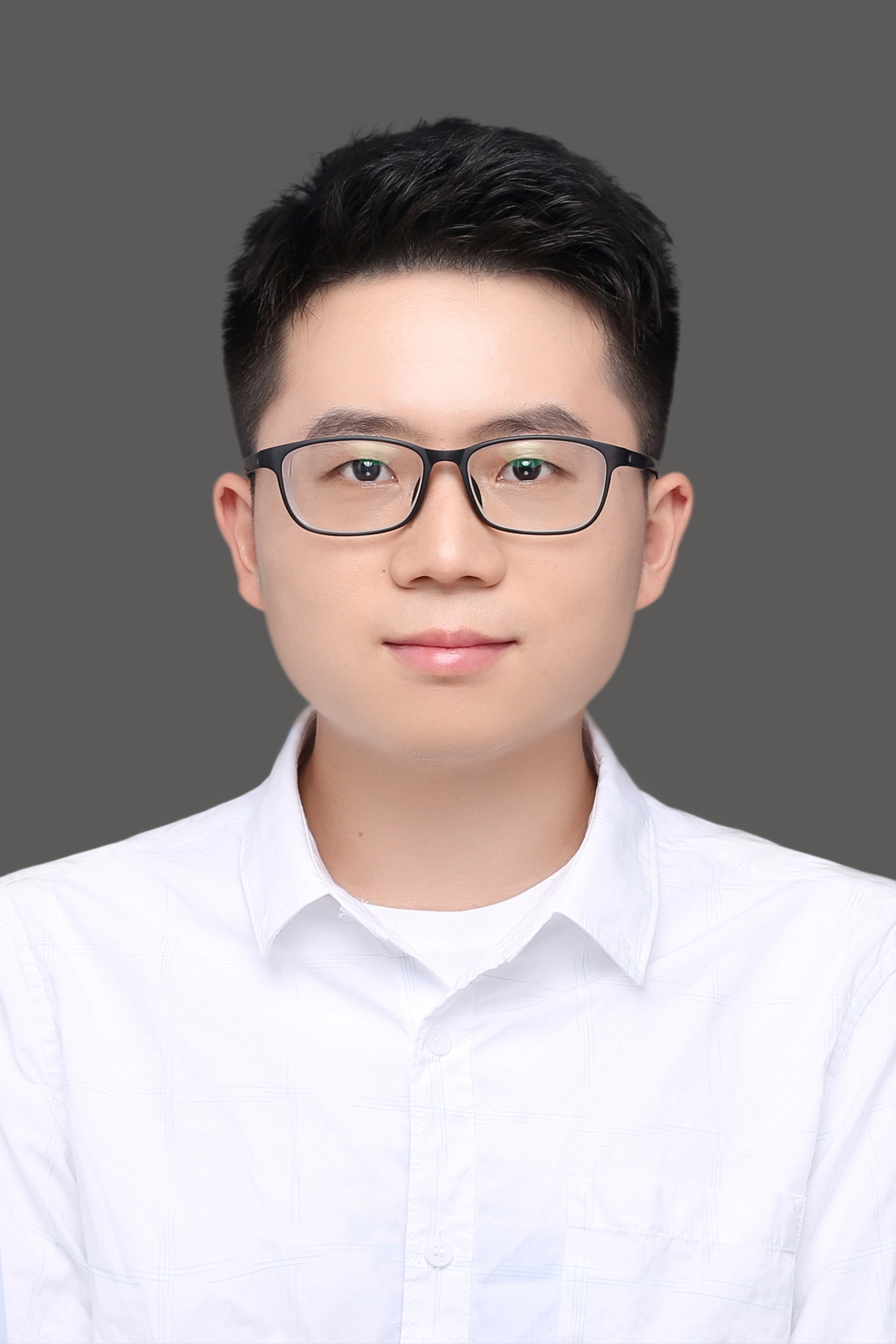}}]{Xiangli Xiao}  received the B.E. degree in communication engineering from the College of Electronic and Information Engineering, Southwest University, Chongqing, China, in Jun. 2020. He is currently working troward the Ph.D. degree in the College of Computer Science and Technology, Nanjing University of Aeronautics and Astronautics, Nanjing, China. His current research interests include multimedia security, digital watermarking, blockchain, and cloud computing security.
\end{IEEEbiography}

\begin{IEEEbiography}[{\includegraphics[width=1in,height=1.25in,clip,keepaspectratio]{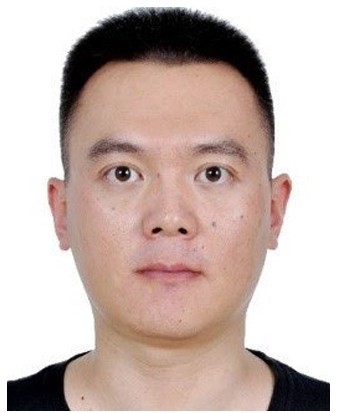}}]{Hua Zhang} received the Ph.D. degree in com puter science from the School of Computer Science	and Technology, Tianjin University, Tianjin, China, in 2015.  He is currently an Associate Professor with the Institute of Information Engineering, Chinese Academy of Sciences. His current research interests include computer vision, multimedia understanding, and machine learning.
\end{IEEEbiography}

\begin{IEEEbiography}[{\includegraphics[width=1in,height=1.25in,clip,keepaspectratio]{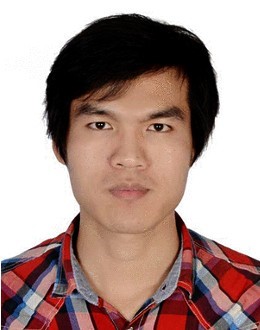}}]{Zhongyun Hua}  received the B.S. degree from Chongqing University, Chongqing, China, in 2011, and the M.S. and Ph.D. degrees from the University of Macau, Macau, China, in 2013 and 2016, respectively. He is currently an Associate Professor with the School of Computer Science and Technology, Harbin Institute of Technology, Shenzhen, China. His current research interests include chaotic systems, multimedia security, and secure cloud computing. He has been recognized as a “Highly Cited Researcher 2022”.
\end{IEEEbiography}






\end{document}